\documentclass[10pt, a4paper]{article}

\usepackage[final]{lrec-coling2024} % this is the new style

\usepackage{amsmath}
\usepackage{booktabs}
\usepackage{enumitem}

\usepackage{arydshln}
\usepackage{lscape}

\title{Is Modularity Transferable?\\ A Case Study through the Lens of Knowledge Distillation}
\name{Mateusz Klimaszewski$^1$, Piotr Andruszkiewicz$^1$, Alexandra Birch$^2$}
\address{
  Institute of Computer Science, Warsaw University of Technology$^1$ \\
  School of Informatics, University of Edinburgh$^2$ \\
  \texttt{mateusz.klimaszewski.dokt@pw.edu.pl} \\}
  
\abstract{
The rise of Modular Deep Learning showcases its potential in various Natural Language Processing applications.
Parameter-efficient fine-tuning (PEFT) modularity has been shown to work for various use cases, from domain adaptation to multilingual setups. However, all this work covers the case where the modular components are trained and deployed within one single Pre-trained Language Model (PLM). This model-specific setup is a substantial limitation on the very modularity that modular architectures are trying to achieve.
We ask whether current modular approaches are transferable between models and whether we can transfer the modules from more robust and larger PLMs to smaller ones.
In this work, we aim to fill this gap via a lens of Knowledge Distillation, commonly used for model compression, and present an extremely straightforward approach to transferring pre-trained, task-specific PEFT modules between same-family PLMs. Moreover, we propose a method that allows the transfer of modules between incompatible PLMs without any change in the inference complexity. The experiments on Named Entity Recognition, Natural Language Inference, and Paraphrase Identification tasks over multiple languages and PEFT methods showcase the initial potential of transferable modularity.
 \\ \newline \Keywords{Modular Deep Learning, Parameter-Efficient Fine-tuning, Pre-trained Language Models}}

\begin{document}

\maketitleabstract

\section{Introduction}
Modular Deep Learning has recently garnered interest as a paradigm that builds upon the idea that a model is a combination of modules with control of the information flow. This paradigm allows for the transfer of learning from one task or language to another, compositionality of the modules and parameter efficiency \citep{pfeiffer2023modulardeeplearning}. For instance, modules allow for efficient (parameter-wise) fine-tuning of Large Language Models \citep{hu2022lora}, enhance task-level generalisation \citep{ponti-etal-2023-combining}, improve multilingual models \citep{bapna-firat-2019-simple}, offer zero-shot capabilities \citep{philip-etal-2020-monolingual} and enable cross-lingual \citep{ansell-etal-2022-composable} or cross-domain \citep{klimaszewski23} knowledge transfer. Furthermore, repositories that store pre-trained modules like AdapterHub \citep{pfeiffer2020AdapterHub} promote the re-usability of previously trained components to new use cases.

\begin{figure}[t!]
\begin{center}
\includegraphics[width=\linewidth]{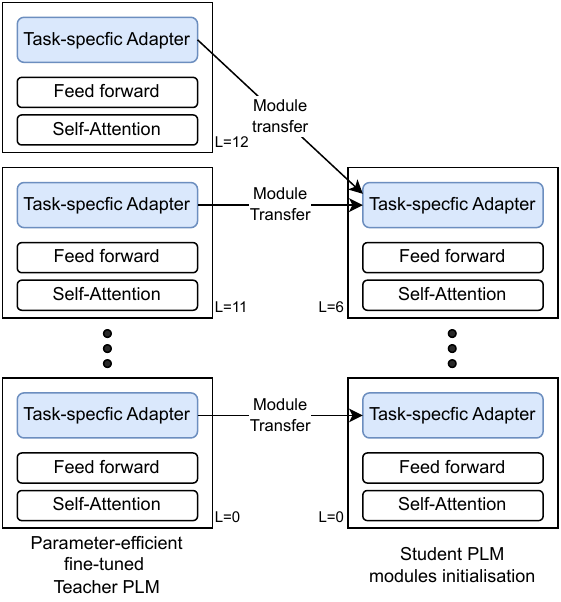} 
\caption{The most straightforward case of transferable modularity. The teacher model is first trained on a task using PEFT, e.g. Adapters, and then the student PEFT modules, prior to fine-tuning, are initialised with the teacher weights.
}
\label{fig:teaser}
\end{center}
\end{figure}

The current modular approaches primarily focus on transferring knowledge to new languages, domains, or tasks. However, prior research assumes that the base model remains constant and overlooks the concept of \textit{transferable modularity}, which entails the potential to transfer modules between different models. From a practical perspective, the effective utilisation of the \textit{transferable modularity property} can reduce the computational burden, especially given the ongoing scaling of Large Language Models \citep{NEURIPS2020_1457c0d6,Touvron2023Llama2O}, allowing for broader re-usability. Moreover, transferring modules from larger to smaller models can significantly enhance knowledge transfer. And finally, even the term ``modularity'' inherently implies the transfer property, suggesting that modular approaches should not be limited to a specific base model.

In this work, we aim to initialise the research objective of \textit{transferable modularity}. We focus on a setup similar to Knowledge Distillation (KD) \citep{Hinton2015DistillingTK}, i.e. where we have two differently sized PLMs (through the paper, we adopt the KD nomenclature, where the bigger model is called a teacher and the smaller - student). Unlike KD, we do not want to use the teacher model's output directly to train a student but use exclusively its fine-tuned PEFT modules.

We show that given matching PLMs (e.g. BERT \cite{devlin-etal-2019-bert} and DistilBERT \citep{distilbert}), it is possible to use pre-trained modules like Adapters \citep{pmlr-v97-houlsby19a,pfeiffer-etal-2021-adapterfusion} or LoRA \citep{hu2022lora} as a better starting point for parameter-efficient (PE) fine-tuning of a smaller student PLM (see Figure \ref{fig:teaser}). Moreover, we investigate a more challenging setup where the models are \textit{incompatible}, i.e., have different internal dimensionality,  and adapt modules via the proposed pruning and alignment method (without inference-time overhead).

To summarise, our contributions are as follows\footnote{Code available at \url{https://github.com/mklimasz/transferable-modularity}}:
\begin{itemize}
    \item We define the property of transferable modularity.
    \item We investigate transferable modularity in matching and incompatible PLMs, proposing a pruning and alignment method for the latter.
\end{itemize}

\section{Transferable Modularity}

\begin{figure}[t]
\begin{center}
\includegraphics[]{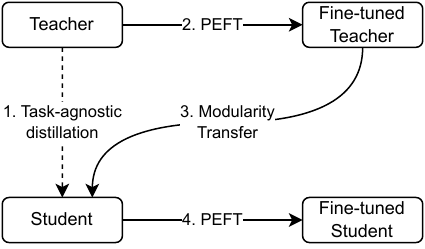} 
\caption{The schema of transferable modularity experiment. We investigate setups where the teacher-student pair result from task-agnostic distillation or are independently trained models.}
\label{fig:kd}
\end{center}
\end{figure}

The high-level idea of our study is presented in Figure \ref{fig:kd}. Given a pair of PLMs, a teacher and a student, we aim to transfer the parameter-efficient (PE) modules from the teacher to the student. First, we use a PEFT technique to train the teacher and its PE modules. Then, we ``move'' the modules from the teacher and insert them into the student, followed by PEFT of the student. This approach means that PE modules of the student have non-random prior initialisation during training.

We consider two setups: (1) matching PLMs and (2) incompatible PLMs. The former uses a shallow version of a teacher with task-agnostic distillation as a student \cite{kim-hassan-2020-fastformers}. This case means that the models represent the same knowledge, have the same hidden dimensionality, and the only difference is the depth of the model. The latter represents a generalised version, where the models are differently parameterised (in terms of latent space size) and they are independently trained. We propose a parameter-free, sample-based pruning and alignment method to answer dimensionality mismatch.

\subsection{Pruning and Alignment}

\begin{figure}[t]
\begin{center}
\includegraphics[width=\linewidth]{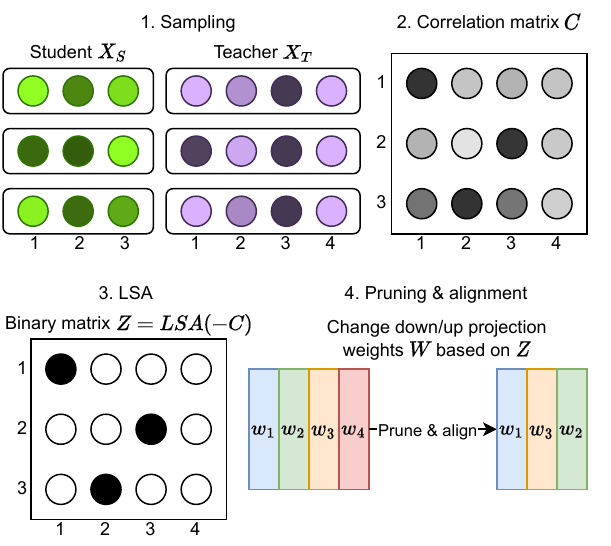} 
\caption{Toy example of adapting the PEFT modules in the case of mismatched dimensionality. Based on the sampled embeddings (1.), correlation matrix $C$ is calculated (2.) and reduced via $LSA$ to a binary matrix $Z$ (3.). In the last step (4.), the pruning and alignment mapping function (derived from $Z$) is applied to down/up projection matrices of LoRA/Adapter modules and match dimensions.}
\label{fig:alg}
\end{center}
\end{figure}

In the case of incompatible PLMs, a dimensionality mismatch problem causes two main issues for transferable modularity. First, the module expects different (higher) dimensionality. Additionally, there exists an alignment discrepancy between the latent spaces of the two models, i.e. if the models have learned the same features, we do not have any guarantee of their placement in the latent space - their indices. 

A crucial element of a successful Knowledge Distillation framework is the computational overhead; therefore, we propose an offline, parameter-free solution that does not change the final student model. The method presented in Figure \ref{fig:alg} consists of four phrases:

\begin{itemize}
    \item sampling
    \item calculating correlation
    \item solving linear sum assignment (LSA) problem 
    \item pruning \& alignment
\end{itemize}
\setlist{}

At first, we sample matching embeddings that would be an input to a PEFT module (we denote the set of embeddings $X_s$ for student and $X_t$ for teacher with $x_s\in X_s$ and $x_t \in X_t$). We store embeddings per layer $l$ (for clarity, we omit the notation of the layer).

In the next step, we establish a correlation matrix between latent spaces. We calculate Pearson's correlation coefficient matrix $C$. $C_{ij}$ is a correlation between the $i$ dimension of a $x_s$ and the $j$ dimension of a $x_t$ embedding.

Given the correlation matrix, we attempt to find the best possible alignment. We define the problem as a linear sum assignment (LSA) \citep{7738348} to establish the optimal mapping. As LSA calculates the minimum cost assignment, we use $-C$ as an input to the LSA algorithm. The algorithm produces a binary matrix $Z$ where $Z_{ij}=1$ means that the $i$ index of $X_s$ is mapped to $j$ of $X_t$.

\begin{equation*}
    min \sum_i \sum_j (-C_{ij})Z_{ij}
\end{equation*}

Finally, using the calculated assignment indices, we remove not-mapped weights from both down/up projection weights $W$ of PEFT modules.

\section{Experiments}
\subsection{Datasets}
To evaluate our method, we benchmark it on three tasks: Named Entity Recognition (NER), Paraphrase Identification (PI) and Natural Language Inference (NLI) using multilingual datasets: WikiNeural \citep{tedeschi-etal-2021-wikineural-combined}, PAWS-X \citep{yang-etal-2019-paws} and XNLI \citep{conneau-etal-2018-xnli} covering jointly a set of over 20 languages\footnote{Arabic, Bulgarian, Chinese, Dutch, English, French, German, Hindi, Italian, Japanese, Korean, Greek, Polish, Portuguese, Russian, Spanish, Swahili, Thai, Turkish, Urdu, Vietnamese}.

\subsection{Training Setup}
We fine-tune multilingual models for each language/task pair using two PEFT methods: Adapter (architecture of \citet{pfeiffer-etal-2021-adapterfusion}, bottleneck size of $96$) and LoRA (rank $8$). We provide the training setup details for each dataset in Appendix \ref{sec:training}.

For teacher-student pairs, we define two configurations:
\begin{itemize}
    \item \textit{matching}: multilingual BERT (mBERT\footnote{\texttt{bert-base-multilingual-cased}}, teacher) -- multilingual DistilBERT (D'mBERT\footnote{\texttt{distilbert-base-multilingual-cased}}, student)
    \item \textit{incompatible}: XLM-RoBERTa Large (XLM-R$_{\text{LARGE}}$\footnote{\texttt{xlm-roberta-large}}, teacher) -- XLM-RoBERTa Base (XLM-R$_{\text{BASE}}$\footnote{\texttt{xlm-roberta-base}}, student) \citep{conneau-etal-2020-unsupervised}
\end{itemize}
We report the relevant hyper-parameters of the models in Table \ref{tab:modelsizes}. As the models have mismatched layer counts, we test two approaches: skip modules (denoted SKIP, e.g., transfer every second module) or average them (denoted AVG, e.g., average the first and second layer's teacher module and transfer to the first module of a student).

\begin{table}[]
    \centering
    \begin{tabular}{lccc}
    \toprule
    Model & Params & Layers & Hidden dim \\
    \midrule
    D'mBERT & $135M$ & $6$ & $768$ \\
    mBERT & $178M$ & $12$ & $768$ \\ \hdashline
    XLM-R$_{\text{BASE}}$ & $278M$ & $12$ & $768$ \\
    XLM-R$_{\text{LARGE}}$ & $560M$ & $24$ & $1024$ \\
    \bottomrule
    \end{tabular}
    \caption{Parameters, layer count and hidden dimension size of the evaluated models.}
    \label{tab:modelsizes}
\end{table}

\subsection{Baselines and Metrics}
For both \textit{matching} and \textit{incompatible} experiments, we define the following structure. As an upper bound of our evaluation, we provide the teacher results after PEFT (Step 2 in Figure \ref{fig:kd}). The baseline is a parameter-efficient fine-tuned student with default modules initialisation (i.e. omitting Step 3 in Figure \ref{fig:kd}).

We report F1 for NER and Accuracy for PI and NLI tasks with an average score over all languages in Section \ref{sec:results}. The detailed per-language results are provided in Appendix \ref{sec:full_results}.

\section{Results and Discussion}
\label{sec:results}

\begin{table*}
\begin{center}
\begin{tabular}{lcccccc}
\toprule
 & \multicolumn{2}{c}{NER (F1)} & \multicolumn{2}{c}{PI (Acc)} & \multicolumn{2}{c}{NLI (Acc)} \\
 & AVG & REL & AVG & REL & AVG & REL \\
\midrule
& \multicolumn{6}{c}{Adapter} \\
Teacher & $95,35$ && $82,60$ && $67,98$ & \\ \hdashline
Student  & $92,94$ & $-2,41$ & $71,32$ & $-11,28$ & $62,12$ & $-5,86$ \\
TM-Student$_{\text{AVG}}$ & $93,02$ & $-2,32$ & $72,96$ & $-9,64$ & $62,33$ & $-5,65$ \\
TM-Student$_{\text{SKIP}}$ & $93,45$ & $-1,90$ & $75,11$ & $-7,49$ & $63,01$ & $-4,97$ \\
& \multicolumn{6}{c}{LoRA} \\
Teacher & $93,27$ && $74,68$ && $63,00$ & \\ \hdashline
Student & $90,09$ & $-3,18$ & $65,80$ & $-8,88$ & $60,56$ & $-2,43$ \\
TM-Student$_{\text{AVG}}$ & $90,63$ & $-2,64$ & $68,52$ & $-6,16$ & $60,53$ & $-2,47$ \\
TM-Student$_{\text{SKIP}}$ & $90,80$ & $-2,47$ & $70,69$ & $-3,99$ & $60,52$ & $-2,47$ \\
\bottomrule
\end{tabular}
\caption{Results of the \textit{matching} PLMs experiment. We report an average score (F1 or Accuracy) over all the datasets' languages and a relative performance gap to the teacher model.}
\label{tab:matching}
\end{center}
\end{table*}

\begin{table*}
\begin{center}
\begin{tabular}{lcccc}
\toprule
 & \multicolumn{2}{c}{NER (F1)} & \multicolumn{2}{c}{PI (Acc)} \\
& AVG & REL & AVG & REL \\
\midrule
& \multicolumn{4}{c}{Adapter} \\
Teacher & $95,34$ && $88,81$ & \\ \hdashline
Student & $93,30$ & $-2,04$ & $84,12$ & $-4,69$ \\
TM-Student$_{\text{SKIP}}$ & $93,34$ & $-2,00$ & $84,27$ & $-4,54$ \\
& \multicolumn{4}{c}{LoRA} \\
Teacher & $93,64$ && $87,03$ &\\ \hdashline
Student & $90,83$ & $-2,82$ & $78,72$ & $-8,31$ \\
TM-Student$_{\text{SKIP}}$ & $90,84$ & $-2,80$ & $78,64$ & $-8,39$ \\
\bottomrule
\end{tabular}
\caption{Results of the \textit{incompatible} PLMs experiment.}
\label{tab:incompatible}
\end{center}
\end{table*}

\subsection{Matching Models}
Table \ref{tab:matching} presents the results of the \textit{matching} experiments. The prefix TM denotes the transfer modularity experiments. The initialisation of the modules transferred from the teacher PLM improved over a default initialisation on average in all the evaluated tasks. Moreover, the SKIP method presents consistency; the difference compared to the baseline was positive across most tasks and languages ($88,7\%$ cases). While at times the improvement was marginal ($+0.02$ gain in Swahili in NLI task), in most cases, as averages indicate, our approach significantly closes the gap to the teacher model (e.g. $+4$ point improvement in Korean on PAWS-X datasets using Adapter or over $+2$ in Spanish LoRA on XNLI). SKIP struggles to outperform the baseline exclusively on XNLI when using LoRA. The results are on par; however, even the teacher models struggle with the task, and the knowledge that can be transferred is relatively limited.

The SKIP outperforms AVG across all the experiments. Considering the results and the findings of \citet{10.1145/3357384.3358028} indicating that the Transformer-based models have internal modularity and each layer has its own defined task, we hypothesise that the averaging might not reflect these phenomena. Therefore, in the \textit{incompatible} experiment, we evaluated just the SKIP method.

\subsection{Incompatible Models}
We present the results of the evaluation in Table \ref{tab:incompatible}. In the case of non-distilled PLMs, the TM method does not significantly outperform the baseline. The changes are uneven; while the transfer shows improvement up to almost $+2$ points in Korean PAWS-X, it can also decrease the performance as in French PAWS-X, losing $-1.05$.

The disparity between \textit{matching} and \textit{incompatible} experiments can be attributed to alignment challenges. Models subjected to distillation exhibit reliable alignment, thanks to the inclusion of an auxiliary loss term such as the cosine embedding loss \citep{distilbert} in the task-agnostic distillation process. In contrast, the correlation-based method encounters difficulties when dealing with models of greater depth. Notably, the LSA algorithm yields lower scores for deeper layers. Considering the different representations required for each language and task pair, this outcome implies that independently trained models require more robust alignment techniques to ensure consistent modularity transfer across all encoded features.

\section{Conclusions}

In this work, we present a case study of transferable modularity property. We evaluate current modular techniques in two scenarios: (1) \textit{matching}, where a student is a shallow, task-agnostic distillation of the teacher and (2) \textit{incompatible}, where a student is independently trained, a shallower model with mismatched internal dimensionality.

The results show that the current modular approach can be transferable as the modules from a matching teacher improve the PEFT of a student model. However, when a student is not distilled from the teacher, the evaluated techniques are inconsistent under the transfer condition, showing the limitation of the current modular methods. We hope this study will inspire future work on modular techniques to consider the transferable modularity property under a more challenging incompatible models scenario.

\section{Acknowledgements}

Part of this work was funded from the European Union's Horizon Europe Research and Innovation program under Grant Agreement No 101070631 and from the UK Research and Innovation (UKRI) under the UK government’s Horizon Europe funding guarantee (Grant No 10039436).

The computations in this work were performed on Poznań Supercomputing and Networking Center and Baskerville. The Baskerville Tier 2 HPC was funded by the EPSRC and UKRI through the World Class Labs scheme (EP/T022221/1) and the Digital Research Infrastructure programme (EP/W032244/1) and is operated by Advanced Research Computing at the University of Birmingham.

\nocite{*}
\section{Bibliographical References}\label{sec:reference}

\bibliographystyle{lrec-coling2024-natbib}
\bibliography{lrec-coling2024-example}

\begin{thebibliography}{23}
\expandafter\ifx\csname natexlab\endcsname\relax\def\natexlab#1{#1}\fi

\bibitem[{Ansell et~al.(2022)Ansell, Ponti, Korhonen, and
  Vuli{\'c}}]{ansell-etal-2022-composable}
Alan Ansell, Edoardo Ponti, Anna Korhonen, and Ivan Vuli{\'c}. 2022.
\newblock \href {https://doi.org/10.18653/v1/2022.acl-long.125} {Composable
  sparse fine-tuning for cross-lingual transfer}.
\newblock In \emph{Proceedings of the 60th Annual Meeting of the Association
  for Computational Linguistics (Volume 1: Long Papers)}, pages 1778--1796,
  Dublin, Ireland. Association for Computational Linguistics.

\bibitem[{Bapna and Firat(2019)}]{bapna-firat-2019-simple}
Ankur Bapna and Orhan Firat. 2019.
\newblock \href {https://doi.org/10.18653/v1/D19-1165} {Simple, scalable
  adaptation for neural machine translation}.
\newblock In \emph{Proceedings of the 2019 Conference on Empirical Methods in
  Natural Language Processing and the 9th International Joint Conference on
  Natural Language Processing (EMNLP-IJCNLP)}, pages 1538--1548, Hong Kong,
  China. Association for Computational Linguistics.

\bibitem[{Brown et~al.(2020)Brown, Mann, Ryder, Subbiah, Kaplan, Dhariwal,
  Neelakantan, Shyam, Sastry, Askell, Agarwal, Herbert-Voss, Krueger, Henighan,
  Child, Ramesh, Ziegler, Wu, Winter, Hesse, Chen, Sigler, Litwin, Gray, Chess,
  Clark, Berner, McCandlish, Radford, Sutskever, and
  Amodei}]{NEURIPS2020_1457c0d6}
Tom Brown, Benjamin Mann, Nick Ryder, Melanie Subbiah, Jared~D Kaplan, Prafulla
  Dhariwal, Arvind Neelakantan, Pranav Shyam, Girish Sastry, Amanda Askell,
  Sandhini Agarwal, Ariel Herbert-Voss, Gretchen Krueger, Tom Henighan, Rewon
  Child, Aditya Ramesh, Daniel Ziegler, Jeffrey Wu, Clemens Winter, Chris
  Hesse, Mark Chen, Eric Sigler, Mateusz Litwin, Scott Gray, Benjamin Chess,
  Jack Clark, Christopher Berner, Sam McCandlish, Alec Radford, Ilya Sutskever,
  and Dario Amodei. 2020.
\newblock \href
  {https://proceedings.neurips.cc/paper_files/paper/2020/file/1457c0d6bfcb4967418bfb8ac142f64a-Paper.pdf}
  {Language models are few-shot learners}.
\newblock In \emph{Advances in Neural Information Processing Systems},
  volume~33, pages 1877--1901. Curran Associates, Inc.

\bibitem[{Conneau et~al.(2020)Conneau, Khandelwal, Goyal, Chaudhary, Wenzek,
  Guzm{\'a}n, Grave, Ott, Zettlemoyer, and
  Stoyanov}]{conneau-etal-2020-unsupervised}
Alexis Conneau, Kartikay Khandelwal, Naman Goyal, Vishrav Chaudhary, Guillaume
  Wenzek, Francisco Guzm{\'a}n, Edouard Grave, Myle Ott, Luke Zettlemoyer, and
  Veselin Stoyanov. 2020.
\newblock \href {https://doi.org/10.18653/v1/2020.acl-main.747} {Unsupervised
  cross-lingual representation learning at scale}.
\newblock In \emph{Proceedings of the 58th Annual Meeting of the Association
  for Computational Linguistics}, pages 8440--8451, Online. Association for
  Computational Linguistics.

\bibitem[{Conneau et~al.(2018)Conneau, Rinott, Lample, Williams, Bowman,
  Schwenk, and Stoyanov}]{conneau-etal-2018-xnli}
Alexis Conneau, Ruty Rinott, Guillaume Lample, Adina Williams, Samuel Bowman,
  Holger Schwenk, and Veselin Stoyanov. 2018.
\newblock \href {https://doi.org/10.18653/v1/D18-1269} {{XNLI}: Evaluating
  cross-lingual sentence representations}.
\newblock In \emph{Proceedings of the 2018 Conference on Empirical Methods in
  Natural Language Processing}, pages 2475--2485, Brussels, Belgium.
  Association for Computational Linguistics.

\bibitem[{Crouse(2016)}]{7738348}
David~F. Crouse. 2016.
\newblock \href {https://doi.org/10.1109/TAES.2016.140952} {On implementing 2d
  rectangular assignment algorithms}.
\newblock \emph{IEEE Transactions on Aerospace and Electronic Systems},
  52(4):1679--1696.

\bibitem[{Devlin et~al.(2019)Devlin, Chang, Lee, and
  Toutanova}]{devlin-etal-2019-bert}
Jacob Devlin, Ming-Wei Chang, Kenton Lee, and Kristina Toutanova. 2019.
\newblock \href {https://doi.org/10.18653/v1/N19-1423} {{BERT}: Pre-training of
  deep bidirectional transformers for language understanding}.
\newblock In \emph{Proceedings of the 2019 Conference of the North {A}merican
  Chapter of the Association for Computational Linguistics: Human Language
  Technologies, Volume 1 (Long and Short Papers)}, pages 4171--4186,
  Minneapolis, Minnesota. Association for Computational Linguistics.

\bibitem[{Hinton et~al.(2015)Hinton, Vinyals, and
  Dean}]{Hinton2015DistillingTK}
Geoffrey~E. Hinton, Oriol Vinyals, and Jeffrey Dean. 2015.
\newblock Distilling the knowledge in a neural network.
\newblock \emph{ArXiv}, abs/1503.02531.

\bibitem[{Houlsby et~al.(2019)Houlsby, Giurgiu, Jastrzebski, Morrone,
  De~Laroussilhe, Gesmundo, Attariyan, and Gelly}]{pmlr-v97-houlsby19a}
Neil Houlsby, Andrei Giurgiu, Stanislaw Jastrzebski, Bruna Morrone, Quentin
  De~Laroussilhe, Andrea Gesmundo, Mona Attariyan, and Sylvain Gelly. 2019.
\newblock \href {https://proceedings.mlr.press/v97/houlsby19a.html}
  {Parameter-efficient transfer learning for {NLP}}.
\newblock In \emph{Proceedings of the 36th International Conference on Machine
  Learning}, volume~97 of \emph{Proceedings of Machine Learning Research},
  pages 2790--2799. PMLR.

\bibitem[{Hu et~al.(2022)Hu, Shen, Wallis, Allen-Zhu, Li, Wang, Wang, and
  Chen}]{hu2022lora}
Edward~J Hu, Yelong Shen, Phillip Wallis, Zeyuan Allen-Zhu, Yuanzhi Li, Shean
  Wang, Lu~Wang, and Weizhu Chen. 2022.
\newblock \href {https://openreview.net/forum?id=nZeVKeeFYf9} {Lo{RA}: Low-rank
  adaptation of large language models}.
\newblock In \emph{International Conference on Learning Representations}.

\bibitem[{Kim and Hassan(2020)}]{kim-hassan-2020-fastformers}
Young~Jin Kim and Hany Hassan. 2020.
\newblock \href {https://doi.org/10.18653/v1/2020.sustainlp-1.20}
  {{F}ast{F}ormers: Highly efficient transformer models for natural language
  understanding}.
\newblock In \emph{Proceedings of SustaiNLP: Workshop on Simple and Efficient
  Natural Language Processing}, pages 149--158, Online. Association for
  Computational Linguistics.

\bibitem[{Klimaszewski et~al.(2023)Klimaszewski, Belligoli, Kumar, and
  Stergiadis}]{klimaszewski23}
Mateusz Klimaszewski, Zeno Belligoli, Satendra Kumar, and Emmanouil Stergiadis.
  2023.
\newblock \href {https://doi.org/10.3233/FAIA230404} {Gated adapters for
  multi-domain neural machine translation}.
\newblock In \emph{{ECAI} 2023 - 26th European Conference on Artificial
  Intelligence}, volume 372 of \emph{Frontiers in Artificial Intelligence and
  Applications}, pages 1264--1271. {IOS} Press.

\bibitem[{Pfeiffer et~al.(2021)Pfeiffer, Kamath, R{\"u}ckl{\'e}, Cho, and
  Gurevych}]{pfeiffer-etal-2021-adapterfusion}
Jonas Pfeiffer, Aishwarya Kamath, Andreas R{\"u}ckl{\'e}, Kyunghyun Cho, and
  Iryna Gurevych. 2021.
\newblock \href {https://doi.org/10.18653/v1/2021.eacl-main.39}
  {{A}dapter{F}usion: Non-destructive task composition for transfer learning}.
\newblock In \emph{Proceedings of the 16th Conference of the European Chapter
  of the Association for Computational Linguistics: Main Volume}, pages
  487--503, Online. Association for Computational Linguistics.

\bibitem[{Pfeiffer et~al.(2020{\natexlab{a}})Pfeiffer, R\"uckl\'{e}, Poth,
  Kamath, Vuli\'{c}, Ruder, Cho, and Gurevych}]{pfeiffer2020AdapterHub}
Jonas Pfeiffer, Andreas R\"uckl\'{e}, Clifton Poth, Aishwarya Kamath, Ivan
  Vuli\'{c}, Sebastian Ruder, Kyunghyun Cho, and Iryna Gurevych.
  2020{\natexlab{a}}.
\newblock \href {https://www.aclweb.org/anthology/2020.emnlp-demos.7}
  {Adapterhub: A framework for adapting transformers}.
\newblock In \emph{Proceedings of the 2020 Conference on Empirical Methods in
  Natural Language Processing (EMNLP 2020): Systems Demonstrations}, pages
  46--54, Online. Association for Computational Linguistics.

\bibitem[{Pfeiffer et~al.(2020{\natexlab{b}})Pfeiffer, R{\"u}ckl{\'e}, Poth,
  Kamath, Vuli{\'c}, Ruder, Cho, and Gurevych}]{pfeiffer-etal-2020-adapterhub}
Jonas Pfeiffer, Andreas R{\"u}ckl{\'e}, Clifton Poth, Aishwarya Kamath, Ivan
  Vuli{\'c}, Sebastian Ruder, Kyunghyun Cho, and Iryna Gurevych.
  2020{\natexlab{b}}.
\newblock \href {https://doi.org/10.18653/v1/2020.emnlp-demos.7}
  {{A}dapter{H}ub: A framework for adapting transformers}.
\newblock In \emph{Proceedings of the 2020 Conference on Empirical Methods in
  Natural Language Processing: System Demonstrations}, pages 46--54, Online.
  Association for Computational Linguistics.

\bibitem[{Pfeiffer et~al.(2023)Pfeiffer, Ruder, Vuli{\'c}, and
  Ponti}]{pfeiffer2023modulardeeplearning}
Jonas Pfeiffer, Sebastian Ruder, Ivan Vuli{\'c}, and Edoardo Ponti. 2023.
\newblock \href {https://openreview.net/forum?id=z9EkXfvxta} {Modular deep
  learning}.
\newblock \emph{Transactions on Machine Learning Research}.
\newblock Survey Certification.

\bibitem[{Philip et~al.(2020)Philip, Berard, Gall{\'e}, and
  Besacier}]{philip-etal-2020-monolingual}
Jerin Philip, Alexandre Berard, Matthias Gall{\'e}, and Laurent Besacier. 2020.
\newblock \href {https://doi.org/10.18653/v1/2020.emnlp-main.361} {Monolingual
  adapters for zero-shot neural machine translation}.
\newblock In \emph{Proceedings of the 2020 Conference on Empirical Methods in
  Natural Language Processing (EMNLP)}, pages 4465--4470, Online. Association
  for Computational Linguistics.

\bibitem[{Ponti et~al.(2023)Ponti, Sordoni, Bengio, and
  Reddy}]{ponti-etal-2023-combining}
Edoardo~Maria Ponti, Alessandro Sordoni, Yoshua Bengio, and Siva Reddy. 2023.
\newblock \href {https://doi.org/10.18653/v1/2023.eacl-main.49} {Combining
  parameter-efficient modules for task-level generalisation}.
\newblock In \emph{Proceedings of the 17th Conference of the European Chapter
  of the Association for Computational Linguistics}, pages 687--702, Dubrovnik,
  Croatia. Association for Computational Linguistics.

\bibitem[{Sanh et~al.(2019)Sanh, Debut, Chaumond, and Wolf}]{distilbert}
Victor Sanh, Lysandre Debut, Julien Chaumond, and Thomas Wolf. 2019.
\newblock \href {http://arxiv.org/abs/1910.01108} {Distilbert, a distilled
  version of {BERT:} smaller, faster, cheaper and lighter}.
\newblock \emph{CoRR}, abs/1910.01108.

\bibitem[{Tedeschi et~al.(2021)Tedeschi, Maiorca, Campolungo, Cecconi, and
  Navigli}]{tedeschi-etal-2021-wikineural-combined}
Simone Tedeschi, Valentino Maiorca, Niccol{\`o} Campolungo, Francesco Cecconi,
  and Roberto Navigli. 2021.
\newblock \href {https://doi.org/10.18653/v1/2021.findings-emnlp.215}
  {{W}iki{NE}u{R}al: {C}ombined neural and knowledge-based silver data creation
  for multilingual {NER}}.
\newblock In \emph{Findings of the Association for Computational Linguistics:
  EMNLP 2021}, pages 2521--2533, Punta Cana, Dominican Republic. Association
  for Computational Linguistics.

\bibitem[{Touvron et~al.(2023)Touvron, Martin, Stone, Albert, Almahairi,
  Babaei, Bashlykov, Batra, Bhargava, Bhosale, Bikel, Blecher, Ferrer, Chen,
  Cucurull, Esiobu, Fernandes, Fu, Fu, Fuller, Gao, Goswami, Goyal, Hartshorn,
  Hosseini, Hou, Inan, Kardas, Kerkez, Khabsa, Kloumann, Korenev, Koura,
  Lachaux, Lavril, Lee, Liskovich, Lu, Mao, Martinet, Mihaylov, Mishra,
  Molybog, Nie, Poulton, Reizenstein, Rungta, Saladi, Schelten, Silva, Smith,
  Subramanian, Tan, Tang, Taylor, Williams, Kuan, Xu, Yan, Zarov, Zhang, Fan,
  Kambadur, Narang, Rodriguez, Stojnic, Edunov, and
  Scialom}]{Touvron2023Llama2O}
Hugo Touvron, Louis Martin, Kevin~R. Stone, Peter Albert, Amjad Almahairi,
  Yasmine Babaei, Nikolay Bashlykov, Soumya Batra, Prajjwal Bhargava, Shruti
  Bhosale, Daniel~M. Bikel, Lukas Blecher, Cristian~Cant{\'o}n Ferrer, Moya
  Chen, Guillem Cucurull, David Esiobu, Jude Fernandes, Jeremy Fu, Wenyin Fu,
  Brian Fuller, Cynthia Gao, Vedanuj Goswami, Naman Goyal, Anthony~S.
  Hartshorn, Saghar Hosseini, Rui Hou, Hakan Inan, Marcin Kardas, Viktor
  Kerkez, Madian Khabsa, Isabel~M. Kloumann, A.~V. Korenev, Punit~Singh Koura,
  Marie-Anne Lachaux, Thibaut Lavril, Jenya Lee, Diana Liskovich, Yinghai Lu,
  Yuning Mao, Xavier Martinet, Todor Mihaylov, Pushkar Mishra, Igor Molybog,
  Yixin Nie, Andrew Poulton, Jeremy Reizenstein, Rashi Rungta, Kalyan Saladi,
  Alan Schelten, Ruan Silva, Eric~Michael Smith, R.~Subramanian, Xia Tan, Binh
  Tang, Ross Taylor, Adina Williams, Jian~Xiang Kuan, Puxin Xu, Zhengxu Yan,
  Iliyan Zarov, Yuchen Zhang, Angela Fan, Melanie Kambadur, Sharan Narang,
  Aurelien Rodriguez, Robert Stojnic, Sergey Edunov, and Thomas Scialom. 2023.
\newblock \href {https://api.semanticscholar.org/CorpusID:259950998} {Llama 2:
  Open foundation and fine-tuned chat models}.
\newblock \emph{ArXiv}, abs/2307.09288.

\bibitem[{van Aken et~al.(2019)van Aken, Winter, L\"{o}ser, and
  Gers}]{10.1145/3357384.3358028}
Betty van Aken, Benjamin Winter, Alexander L\"{o}ser, and Felix~A. Gers. 2019.
\newblock \href {https://doi.org/10.1145/3357384.3358028} {How does bert answer
  questions? a layer-wise analysis of transformer representations}.
\newblock In \emph{Proceedings of the 28th ACM International Conference on
  Information and Knowledge Management}, CIKM '19, page 1823–1832, New York,
  NY, USA. Association for Computing Machinery.

\bibitem[{Yang et~al.(2019)Yang, Zhang, Tar, and
  Baldridge}]{yang-etal-2019-paws}
Yinfei Yang, Yuan Zhang, Chris Tar, and Jason Baldridge. 2019.
\newblock \href {https://doi.org/10.18653/v1/D19-1382} {{PAWS}-{X}: A
  cross-lingual adversarial dataset for paraphrase identification}.
\newblock In \emph{Proceedings of the 2019 Conference on Empirical Methods in
  Natural Language Processing and the 9th International Joint Conference on
  Natural Language Processing (EMNLP-IJCNLP)}, pages 3687--3692, Hong Kong,
  China. Association for Computational Linguistics.

\end{thebibliography}

% \section{Language Resource References}
% \label{lr:ref}
% \bibliographystylelanguageresource{lrec-coling2024-natbib}
% \bibliographylanguageresource{languageresource}

\appendix

\section{Experimental Setup}
\label{sec:training}
We use the AdapterHub library \cite{pfeiffer-etal-2020-adapterhub} for all our experiments. We train all our models using a single GPU with a batch size of $64$ and a learning rate of 1e--5 for $10$ epochs for NER \& NLI tasks and $30$ epochs for the PI task. We choose the final checkpoint based on validation dataset performance.

For PEFT hyper-parameters, we set the bottleneck size to $96$ for Adapter modules and a rank of $8$ for LoRA. We apply LoRA to the query and value self-attention modules.

\section{Per Language Results}
\label{sec:full_results}
In Tables \ref{tab:ner_full}, \ref{tab:pi_full}  and \ref{tab:nli_full}, we expand the results reported in Tables \ref{tab:matching} and \ref{tab:incompatible} and provide the scores for each evaluated language.

\begin{table*}
\begin{center}
\begin{tabular}{lccccccccc}
\toprule
Model & de & en & es & fr & it & nl & pl & pt & ru \\
\midrule
& \multicolumn{9}{c}{\textit{Matching PLMs}} \\
& \multicolumn{9}{c}{Adapter} \\
Teacher & $97,57$ & $92,79$ & $98,08$ & $95,49$ & $94,85$ & $97,74$ & $95,54$ & $95,91$ & $90,15$ \\
\hdashline
Student & $95,34$ & $90,23$ & $95,81$ & $93,23$ & $92,67$ & $95,27$ & $93,73$ & $94,21$ & $85,97$ \\
TM-Student$_{\text{AVG}}$ & $95,53$ & $90,21$ & $95,87$ & $93,26$ & $92,74$ & $95,39$ & $93,88$ & $94,40$ & $85,92$ \\
TM-Student$_{\text{SKIP}}$ & $95,88$ & $90,65$ & $96,24$ & $93,77$ & $93,13$ & $95,87$ & $94,20$ & $94,72$ & $86,57$ \\
& \multicolumn{9}{c}{Lora} \\
Teacher & $95,78$ & $90,49$ & $96,45$ & $93,26$ & $93,13$ & $95,94$ & $93,97$ & $94,41$ & $85,97$ \\
\hdashline
Student & $92,45$ & $87,25$ & $93,55$ & $90,06$ & $89,95$ & $92,85$ & $91,55$ & $92,15$ & $80,96$ \\
TM-Student$_{\text{AVG}}$ & $92,92$ & $87,74$ & $93,94$ & $90,57$ & $90,75$ & $93,24$ & $91,97$ & $92,52$ & $82,03$ \\
TM-Student$_{\text{SKIP}}$ & $93,03$ & $87,99$ & $93,87$ & $90,88$ & $91,04$ & $93,44$ & $92,04$ & $92,61$ & $82,27$ \\
& \multicolumn{9}{c}{\textit{Incompatible PLMs}} \\
& \multicolumn{9}{c}{Adapter} \\
Teacher & $97,36$ & $92,30$ & $97,95$ & $95,61$ & $94,99$ & $97,79$ & $96,15$ & $96,12$ & $89,74$ \\
\hdashline
Student & $95,20$ & $89,92$ & $96,19$ & $93,34$ & $93,06$ & $96,29$ & $94,14$ & $94,56$ & $86,99$ \\
TM-Student$_{\text{SKIP}}$ & $95,30$ & $90,03$ & $96,21$ & $93,40$ & $93,00$ & $96,09$ & $94,19$ & $94,78$ & $87,01$ \\
& \multicolumn{9}{c}{Lora} \\
Teacher & $94,68$ & $89,94$ & $96,19$ & $92,35$ & $92,85$ & $95,67$ & $93,82$ & $94,11$ & $85,09$ \\
\hdashline
Student & $92,21$ & $86,65$ & $92,74$ & $89,40$ & $90,05$ & $93,14$ & $91,61$ & $92,25$ & $81,62$ \\
TM-Student$_{\text{SKIP}}$ & $92,10$ & $86,65$ & $93,00$ & $89,61$ & $90,01$ & $93,01$ & $91,51$ & $92,32$ & $81,69$ \\
\bottomrule
\end{tabular}
\caption{Named Entity Recognition results per language.}
\label{tab:ner_full}
\end{center}
\end{table*}

\begin{table*}
\begin{center}
\begin{tabular}{lccccccc}
\toprule
Model & de & en & es & fr & ja & ko & zh \\
\midrule
& \multicolumn{7}{c}{\textit{Matching PLMs}} \\
& \multicolumn{7}{c}{Adapter} \\
Teacher & $83,60$ & $91,60$ & $85,20$ & $86,90$ & $76,05$ & $75,95$ & $78,90$ \\
\hdashline
Student & $73,30$ & $75,85$ & $72,90$ & $74,65$ & $67,25$ & $65,25$ & $70,05$ \\
TM-Student$_{\text{AVG}}$ & $74,15$ & $82,35$ & $73,35$ & $75,10$ & $67,10$ & $67,40$ & $71,25$ \\
TM-Student$_{\text{SKIP}}$ & $74,50$ & $85,05$ & $77,85$ & $78,15$ & $69,10$ & $69,25$ & $71,85$ \\
& \multicolumn{7}{c}{Lora} \\
Teacher & $75,10$ & $83,30$ & $78,70$ & $77,75$ & $67,85$ & $67,95$ & $72,10$ \\
\hdashline
Student & $70,20$ & $63,45$ & $67,30$ & $70,10$ & $62,80$ & $61,85$ & $64,90$ \\
TM-Student$_{\text{AVG}}$ & $71,95$ & $69,35$ & $69,95$ & $71,85$ & $64,75$ & $64,05$ & $67,75$ \\
TM-Student$_{\text{SKIP}}$ & $72,25$ & $74,50$ & $72,30$ & $74,85$ & $66,70$ & $64,90$ & $69,30$ \\
& \multicolumn{7}{c}{\textit{Incompatible PLMs}} \\
& \multicolumn{7}{c}{Adapter} \\
Teacher & $90,45$ & $94,70$ & $91,20$ & $92,15$ & $82,35$ & $85,00$ & $85,80$ \\
\hdashline
Student & $85,75$ & $92,55$ & $87,25$ & $89,25$ & $77,10$ & $75,65$ & $81,30$ \\
TM-Student$_{\text{SKIP}}$ & $86,15$ & $92,05$ & $88,50$ & $88,20$ & $76,80$ & $77,45$ & $80,75$ \\
& \multicolumn{7}{c}{Lora} \\
Teacher & $89,40$ & $93,80$ & $89,90$ & $89,65$ & $80,95$ & $81,45$ & $84,05$ \\
\hdashline
Student & $80,00$ & $88,05$ & $82,95$ & $83,55$ & $72,55$ & $68,60$ & $75,35$ \\
TM-Student$_{\text{SKIP}}$ & $80,95$ & $88,05$ & $82,10$ & $82,70$ & $71,65$ & $69,80$ & $75,20$ \\
\bottomrule
\end{tabular}
\caption{Paraphrase Identification results per language.}
\label{tab:pi_full}
\end{center}
\end{table*}

\begin{landscape}
\begin{table}
\begin{center}
%\scalebox{0.75}{

%\resizebox{\textwidth}{!}{
\begin{tabular}{lccccccccccccccc}
\toprule
Model & ar & bg & de & el & en & es & fr & hi & ru & sw & th & tr & ur & vi & zh \\
\midrule
& \multicolumn{15}{c}{Adapter} \\
Teacher & $65,53$ & $70,18$ & $70,12$ & $68,08$ & $77,03$ & $73,01$ & $72,00$ & $63,39$ & $69,58$ & $60,12$ & $60,40$ & $67,49$ & $60,40$ & $71,28$ & $71,10$ \\
\hdashline
Student & $60,12$ & $63,47$ & $65,07$ & $63,35$ & $69,62$ & $66,11$ & $65,89$ & $57,33$ & $62,14$ & $56,99$ & $56,43$ & $61,86$ & $55,71$ & $63,67$ & $64,01$ \\
TM-Student$_{\text{AVG}}$ & $60,54$ & $63,57$ & $65,19$ & $63,27$ & $70,38$ & $66,69$ & $66,13$ & $57,54$ & $62,53$ & $56,17$ & $56,39$ & $62,00$ & $56,59$ & $63,33$ & $64,59$ \\
TM-Student$_{\text{SKIP}}$ & $61,16$ & $64,23$ & $65,27$ & $63,49$ & $70,58$ & $68,16$ & $66,53$ & $58,48$ & $63,45$ & $57,01$ & $57,05$ & $62,61$ & $57,47$ & $63,89$ & $65,73$ \\

& \multicolumn{15}{c}{Lora} \\
Teacher & $61,46$ & $63,79$ & $66,83$ & $63,77$ & $70,12$ & $67,84$ & $66,99$ & $59,90$ & $64,47$ & $54,57$ & $55,37$ & $61,60$ & $56,47$ & $65,01$ & $66,77$ \\
\hdashline
Student & $59,20$ & $62,40$ & $63,43$ & $61,96$ & $67,15$ & $64,09$ & $63,87$ & $57,56$ & $60,62$ & $54,87$ & $53,67$ & $59,84$ & $55,97$ & $61,74$ & $62,08$ \\
TM-Student$_{\text{AVG}}$ & $59,20$ & $62,40$ & $63,43$ & $61,96$ & $67,15$ & $64,09$ & $63,87$ & $57,56$ & $60,62$ & $54,87$ & $53,67$ & $59,72$ & $55,77$ & $61,74$ & $61,88$ \\
TM-Student$_{\text{SKIP}}$ & $59,10$ & $62,36$ & $63,75$ & $61,66$ & $67,19$ & $64,11$ & $64,19$ & $57,56$ & $60,94$ & $54,37$ & $53,19$ & $59,76$ & $56,11$ & $61,30$ & $62,26$ \\
\bottomrule
\end{tabular}

\caption{Natural Language Inference results per language for the \textit{matching} PLMs experiment.}
\label{tab:nli_full}
\end{center}
\end{table}
\end{landscape}

\end{document}